\documentclass[11pt,a4paper]{article}
\usepackage[hyperref]{acl2020}
\usepackage{times}
\usepackage{latexsym}
\usepackage{amsmath}
\usepackage{graphicx}
\usepackage{multirow}

\usepackage{microtype}

\aclfinalcopy 

\usepackage{enumitem}
\usepackage{booktabs}
\title{MCQA: Multimodal Co-attention Based Network for Question Answering}
\author{Abhishek Kumar, Trisha Mittal, Dinesh Manocha \\
  Department of Computer Science, University of Maryland \\
  College Park, Maryland 20740, USA \\
  \texttt{\{akumar09, trisha, dm\}@cs.umd.edu}}

\date{}

\begin{document}
\maketitle
\begin{abstract}
We present MCQA, a learning-based algorithm for multimodal question answering. MCQA explicitly fuses and aligns the multimodal input~(i.e. text, audio, and video), which forms the context for the query~(question and answer). Our approach fuses and aligns the question and the answer within this context.  Moreover, we use the notion of co-attention to perform  cross-modal alignment and multimodal context-query alignment. Our context-query alignment module matches the relevant parts of the multimodal context and the query with each other and aligns them to improve the overall performance.    We evaluate the performance of MCQA on Social-IQ, a benchmark dataset for multimodal question answering. We compare the performance of our algorithm with prior methods and observe an accuracy improvement of $4-7$\%.
\end{abstract}
\section{Introduction}

\label{sec:introduction}
Intelligent Question Answering~\cite{woods1978semantics} has seen great progress in the past decade~\cite{zadeh2019social}. Initial attempts in question answering were limited to a single modality of text~\cite{rajpurkar2018know,rajpurkar2016squad,welbl2018constructing,weston2015towards}. There has been a  shift in terms of using multiple modalities that also include audio or video. For instance, recent works have shown good performance in single-image based question answering, which include image and text modalities~\cite{agrawal2017vqa,jang2017tgif,yu2015visual}, and video-based question answering tasks with video, audio and text as the underlying modalities~\cite{tapaswi2016movieqa,kim2018multimodal,lei2018tvqa}. 

Recently, there has been interest in the community to shift to more challenging questions of the nature \textit{`why'} and \textit{`how'}, rather than \textit{`what'} and \textit{`when'} to make the models more intelligent. These questions are harder as they require a sense of causal reasoning. One such recent benchmark in this context is the Social-IQ~(Social Intelligence Queries) dataset~\cite{zadeh2019social}. This dataset contains a set of $7,500$ questions, $52,500$ answers for $1,250$ social in-the-wild videos. This dataset comprises of a diverse set of videos collected from YouTube, is completely unconstrained and  unscripted, and is regarded as a challenging dataset because of the gap between human performance $(95.08\%)$ and the prior methods on this dataset $(64.82\%)$. The average length of the answers of the Social-IQ dataset is longer than the previous datasets by $100\%$. This makes it challenging to develop accurate algorithms for such datasets.   

\textbf{Main Contributions: } Social-IQ is a challenging dataset for video, text, audio input. We successfully demonstrate better results than SOTA systems on other datasets like MovieQA and TVQA when applied to Social-IQ. Given a input video and an input query~(question and answer), we present a novel learning-based algorithm to predict if the answer is correct or not. Our main contributions include:
\begin{enumerate}[noitemsep]
    \item We present MCQA, a multimodal question answering algorithm that includes two main novel components: \textit{Multimodal Fusion and Alignment}, which fuses and aligns the three multimodal inputs~(text, video and audio) to serve as a context to the query and \textit{Multimodal Context-Query Alignment}, which performs cross-alignment of the modalities with the query. 
    \item We propose the use of \textit{Co-Attention}, a concept borrowed from Machine Reading Comprehension literature~\cite{wang2018multi} to perform these alignments. 
    \item We analyze the performance of MCQA and compare it with the existing state-of-the-art QA methods on the Social-IQ dataset. We report an improvement of around $4-7$\% over prior methods. 
\end{enumerate}
\section{Related Work}
\label{sec:relatedwork}
\begin{figure*}[h]
  \centering
  \includegraphics[width=14.1cm,height=4.45cm]{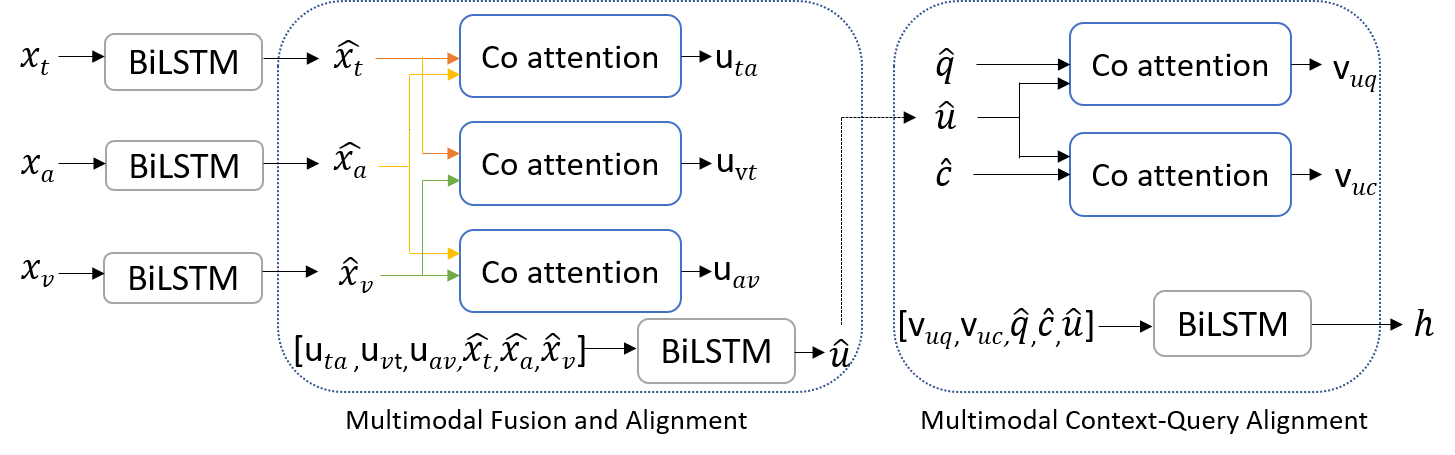}
  \caption{\textbf{MCQA Network Architecture: } The input $(x_{t}, x_{a}, x_{v}, q, c)$ to our network are encoded using \texttt{BiLSTMs}. Our multimodal fusion and alignment component aligns and fuses the multimodal inputs $(x_{t}, x_{a}, x_{v})$ using co-attention and \texttt{BiLSTM} to output the multimodal context $\hat{u}$. Our multimodal context-query alignment module uses co-attention and \texttt{BiLSTM} to soft-align and combine $\hat{u}, \hat{q}, \hat{c}$ to generate the latent representation, $h$. The final answer ($\hat h$) is then computed by applying Eqs $4$ and $5$.}
  \label{fig:network}
  \vspace{-10pt}
\end{figure*}
\textbf{Question Answering Datasets: } Datasets like COCO-QA~\cite{ren2015exploring}, VQA~\cite{antol2015vqa}, FM-IQA~\cite{gao2015you}, and Visual7w~\cite{zhu2016visual7w} are single-image based question answering datasets. MovieQA~\cite{tapaswi2016movieqa} and TVQA~\cite{lei2018tvqa} extend the task of question answering from a single image to videos. All of these datasets have focused on questions like \textit{`what'} and  \textit{`when'}.

\textbf{Multimodal Fusion: }With multiple input modalities, it is important to consider how to fuse different modalities. Fusion methods like Tensor fusion network~\cite{zadeh2017tensor} and memory fusion network~\cite{zadeh2018memory} have been used for multimodal sentiment analysis. These methods use tensor fusion and attention-based memory fusion, respectively.~\citeauthor{lei2018tvqa}~(2018) use a context matching module and BiLSTM to model the inputs in a combined manner for multimodal question answering task. ~\citeauthor{sun2019videobert}~(2019) proposed VideoBERT model to learn the bidirectional joint distributions over text and video data. Our model takes in text, video, and audio features as the input however, VideoBERT only takes in video and text data. It is not evident if VideoBERT can be extended directly to handle audio inputs.

\textbf{Multimodal Context-Query Alignment: } Alignment has been studied extensively for reading comprehension and question answering tasks. For instance, \citeauthor{wang2018multi}~(2018) use a hierarchical attention fusion network to model the query and the context. \citeauthor{xiong2016dynamic}~(2016) use a co-attentive encoder that captures the interactions between the question and the document, and dynamic pointing decoder to answer the questions. \citeauthor{wang2017gated}~(2017) propose a self-matching attention mechanism and use pointer networks.

\section{Our Approach: MCQA}
\label{sec:approach}
In this section, we present the details of our learning-based algorithm(MCQA). In Section~\ref{subsec:overview}, we give an overview of the approach. This is followed by a discussion of co-attention in Section~\ref{subsec:co-attention}. We use this notion for Multimodal Fusion and Alignment and Multimodal Context-Query Alignment. We present these two alignments in Section~\ref{subsec:slignment} and \ref{subsec:query}, respectively.
\subsection{Overview}
\label{subsec:overview}
We present MCQA for the task of multimodal question answering. We directly used the features publicly released by the Social-IQ authors. The input text, audio, and video features were generated using BERT embeddings, COVAREP features, and pretrained DenseNet161 network, respectively. The video frames were sampled at 30fps. We used the CMU-SDK library to align text, audio, and video modalities automatically. Given an input video, with three input modalities, text~($x_{t}$), audio~($x_{a}$), and video~($x_{v}$) of feature length 768, 74, 2208, respectively, we perform Multimodal Fusion and Alignment to obtain the context for the input query. Next, we perform Multimodal Context-Query Alignment to obtain the predicted answer using $\hat{h}$. Figure~\ref{fig:network} highlights the overall network. We tried multiple approaches and different attention mechanisms. Ultimately, co-attention performed the best.
\subsection{Co-attention}
\label{subsec:co-attention}
Our approach is motivated by \cite{wang2018multi}, who use the notion of co-attention for the machine comprehension reading. The co-attention component calculates the shallow semantic similarity between the two inputs. Given two inputs $(\big\{u_{p}^{t}\big\}_{t=1}^{t=T}, \big\{u_{q}^{t}\big\}_{t=1}^{t=T})$, co-attention aligns them by constructing a soft-alignment matrix $S$. Here, $u_{p}$ and $u_{q}$ can be the encoded outputs of any $\texttt{BiLSTM}$.  Each $(i,j)$ entry of the matrix $S$ is the multiplication of the $\texttt{Relu}$ \cite{nair2010rectified} activation for both the inputs.
\begin{equation*}
    S_{i,j} = \texttt{ReLU}(w_{p}u_{p}^{i}) \cdot \texttt{ReLU}(w_{q}u_{q}^{j})
\end{equation*}
We use the attention weights of the matrix $S$ to obtain the attended vectors $\hat{u}_{p}^{t}$ and $\hat{u}_{q}^{t}$. $w_{p}$ and $w_{q}$ are trainable weights which are learned jointly. Each vector $\hat{u}_{p}^{i}$ is the combination of the vectors $u_{q}^{t}$ that are most relevant to $u_{p}^{i}$. Similarly, each $\hat{u}_{q}^{j}$ is the combination of most relevant $u_{p}^{t}$ for $u_{q}^{j}$.
\begin{align*}
    \alpha_{i,:} &= \texttt{softmax}(S_{i,:}), \\
    \hat{u_{p}^{i}} &= \sum_{j}\alpha_{i,j} \cdot u_{q}^{j} \\
    \alpha_{:,j} &= \texttt{softmax}(S_{:,j}) \\
    \hat{u_{q}^{j}} &= \sum_{i}\alpha_{i,j} \cdot u_{p}^{i}
\end{align*}
\begin{equation*}
    u_{pq} = [\hat{u}_{p}, \hat{u}_{q}]
\end{equation*}
 $\alpha_{i,:}$ represents the attention weight on all relevant $u_{q}^{t}$ for $u_{p}^{i}$. Similarly for $u_{p}^{i}$, $\alpha_{:,j}$ represents the attention weight on all relevant $u_{q}^{t}$. The final attended representation $u_{pq}$ is the concatenation of $\hat{u_{p}^{t}}$ and $\hat{u_{q}^{t}}$ which captures the most important parts of $u_{p}^{t}$ and $u_{q}^{t}$ with respect to each other. We use $u_{pq}$ in our network to capture the soft-alignment between $u_{p}$ and $u_{q}$. As an example, given $x_{t}$ and $x_{a}$ as an input to the co-attention component, it will output $u_{ta}$ as shown in Fig~\ref{fig:network}.
\subsection{Multimodal Fusion and Alignment}
\label{subsec:slignment}
The multimodal input from the Social-IQ dataset consists of text ($x_{t}$), audio ($x_{a}$) and video ($x_{v}$) component. These components are encoded using $\texttt{BiLSTMs}$ to capture the contextual information and output $\hat{x}_{t}$, $\hat{x}_{a}$ and $\hat{x}_{v}$.  We utilized all the outputs of BiLSTM and not just the final vector representation.  We use $\texttt{BiLSTMs}$ of dimension $200$, $100$, $250$ for text, audio, and video, respectively. Similarly, the question and the answer are encoded using $\texttt{BiLSTMs}$ to output $\hat{q}$ and $\hat{c}$.
\begin{align*}
    \hat{x}_{t} = \texttt{BiLSTM}&(x_{t}), \quad \hat{x}_{a} = \texttt{BiLSTM}(x_{a}),\\
    &\hat{x}_{v} = \texttt{BiLSTM}(x_{v}),\\
    \hat{q} = \texttt{BiLSTM}&(q), \quad \hat{c} = \texttt{BiLSTM}(c)
\end{align*}
The interactions between the different modalities are captured by the multimodal fusion and alignment components. We use co-attention as described in Section~\ref{subsec:co-attention} to combine different modalities. Next we use a $\texttt{BiLSTM}$ to combine the encoded inputs $(\hat{x}_{t}, \hat{x}_{a}, \hat{x}_{v})$ and the outputs of the modality fusion $(u_{ta}, u_{av} ,u_{vt})$ to obtain the multimodal context representation $\hat{u}$.
\begin{align*}
    u_{ta}&=\texttt{Co-attention}(\hat{x}_{t}, \hat{x}_{a})\\
    u_{av} &= \texttt{Co-attention}(\hat{x}_{a}, \hat{x}_{v}) \\
    u_{vt} &= \texttt{Co-attention}(\hat{x}_{v}, \hat{x}_{t})  
\end{align*}
\begin{equation*}
    \hat{u} = \texttt{BiLSTM}([u_{ta}, u_{av}, u_{vt}, \hat{x}_{t}, \hat{x}_{a}, \hat{x}_{v}])
\end{equation*}
\subsection{Multimodal context - Query alignment}
\label{subsec:query}
The alignment between a context and a question is an important step to locate the answer for the question \cite{wang2018multi}. We use the notion of co-attention  to align the multimodal context and the question to obtain their aligned fused representation $v_{uq}$. Similarly, we align the multimodal context and the answer choice to compute $v_{uc}$.
\begin{align}
    v_{uq} &= \texttt{Co-attention}(\hat{u}, \hat{q}), \\
    v_{uc} &= \texttt{Co-attention}(\hat{u}, \hat{c}), \\
    h &= \texttt{BiLSTM}([v_{uq}, v_{uc}, \hat{q}, \hat{c}, \hat{u}]), \\
    \beta &= \texttt{Softmax}(w_{r}h), \\
    \hat{h} &= \sum_{k}\beta_{k} \cdot h^{k}
\end{align}
We fuse the representation $v_{u,q}$, $v_{u,c}$, $\hat{q}$, $\hat{c}$ and, $\hat{u}$ using a $\texttt{BiLSTM}$. In order to make the final prediction, we obtain $\hat{h}$ using a linear self-alignment\cite{wang2018multi} on $h$ and pass it through a feed-forward neural network. $w_{r}$ is a trainable weight.

\section{Experiments and Results}
\label{sec:experiments-results}
\begin{table}
\centering
\begin{tabular}{lcc}
\toprule \textbf{Models} & \textbf{A2} & \textbf{A4}\\ \midrule
LMN & 61.1\% & 31.8\% \\
FVTA & 60.9\% & 31.0\% \\
E2EMemNet & 62.6\% & 31.5\% \\
MDAM & 60.2\% & 30.7\% \\
MSM & 60.0\% & 29.9\% \\
TFN & 63.2\% & 29.8\% \\
MFN & 62.8\% & 30.8\% \\
Tensor-MFN & 64.8\% & 34.1\% \\ 
\textbf{MCQA} & \textbf{68.8\%} & \textbf{38.3\%} \\
\bottomrule
\end{tabular}
\caption{\label{font-table} \textbf{Accuracy Performance: }We compare the performance of our method with eight prior methods on Social-IQ dataset and observe $4-7\%$ accuracy improvement.}
\label{tab:accuracy}
\vspace{-5pt}
\end{table}
\begin{table}
\centering
\begin{tabular}{lc}
\toprule 
\textbf{Models} & \textbf{A2} \\ \midrule
MCQA w/o fusion and alignment & 66.9\% \\
MCQA w/o context-query alignment & 67.4\%  \\ 
\textbf{MCQA} & \textbf{68.8\%} \\
\bottomrule
\end{tabular}
\caption{\label{font-table} \textbf{Ablation Experiments: }We analyze the contribution of  each of the components proposed by performing ablation experiments and report accuracy numbers by removing some components.}
\label{tab:ablation}
\vspace{-10pt}
\end{table}
\subsection{Training Details}
We train the MCQA with a batch size of 32 for 100 epochs. We use the Adam optimizer \cite{kingma2014adam} with a learning rate of 0.001. All our results were generated on an NVIDIA GeForce GTX 1080 Ti GPU. We performed grid search over the hyperparameter space of number of epochs, learning rate, and dimensions of the BiLSTM.
\subsection{Evaluation Methods}
We analyze the performance of the MCQA by comparing it against Tensor-MFN~\cite{zadeh2019social}, the previous state-of-the-art system on the Social-IQ dataset. We also compare our system with the End2End Multimodal Memory Network~(E2EMMemNet)~\cite{sukhbaatar2015end} and Multimodal Dual Attention Memory~(MDAM)~\cite{kim2018multimodal} which uses self-attention based on visual frames and cross-attention based on question. These systems showed good performance on the MovieQA dataset. We also compared our system with the Layered Memory Network~(LMN)~\cite{wang2018movie}, the winner of ICCV 2017 which uses Static Word Memory Module and the Dynamic Subtitle Memory Module, and with Focal Visual-Text Attention~(FVTA)~\cite{liang2018focal} which proposed Focal Visual-Text attention. These approaches have a strong performance on the MovieQA dataset. We also compare our approach with Multi-stream Memory~(MSM) \cite{lei2018tvqa}, a top-performing baseline for TVQA which encodes all the modalities using recurrent networks. We also compare with Tensor Fusion Network~(TFN)~\cite{zadeh2017tensor} and Memory Fusion Network~(MFN)~\cite{zadeh2018memory}.
\begin{figure}[h]
  \centering
  \includegraphics[width=8cm,height=5cm]{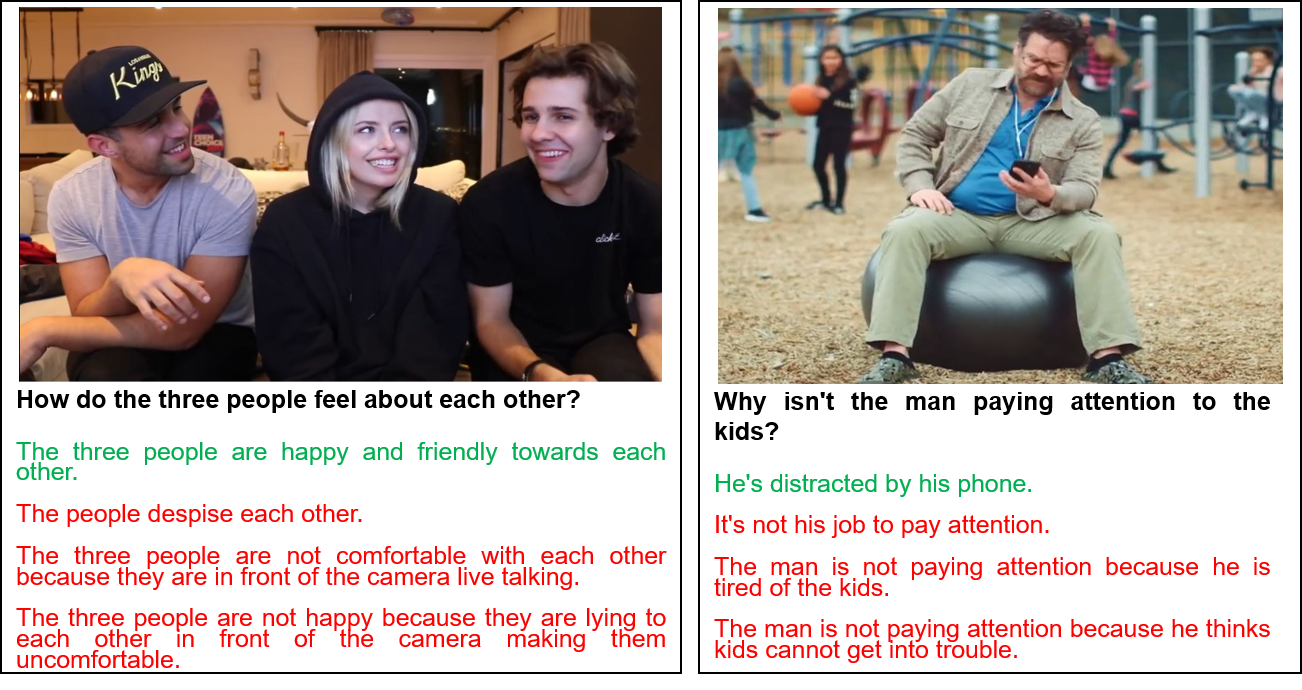}
  \caption{\textbf{Qualitative Results:} Example snippets from the Social-IQ dataset. }
  \label{fig:qualitative}
  \vspace{-15pt}
\end{figure}
\subsection{Analysis and Discussion}
\textbf{Comparison with SOTA Methods: } Table~\ref{tab:accuracy} shows the accuracy performance of the MCQA for A2 and A4 tasks. The A2 task is to select the correct answer from two given answer choices. The A4 task is to select the correct answer from four given answer choices. We observe that MCQA is $4-7$\% better than Tensor-MFN network on the A2 and A4 tasks.\\
\textbf{Qualitative Results:} We show one frame from two videos from the Social-IQ dataset, where our model answers correctly in the Fig \ref{fig:qualitative}. The choice highlighted in green is the correct answer to the answer asked, whereas the red choices indicate the incorrect answers. 
\subsection{Ablation Experiments}
\textbf{MCQA without Multimodal Fusion and Alignment: } We removed the component of the network that is used to fuse and align the input modalities~(text, video and audio) using co-attention. We observe a drop in the performance to $66.9$\%. This can be observed in Row 1 of  Table~\ref{tab:ablation}. We believe this reduction in performance is due to the fact that the network cannot exploit the three modalities to their full potential without the alignment. Our approach builds on the Social-IQ paper, which established the baseline for multimodal question answering. Social-IQ authors observed that the combination of text, audio, and video modality produced the best results, and we use this finding to incorporate all three modalities in all our experiments.\\
\textbf{MCQA without Multimodal Context-Query Alignment: }As shown in Table~\ref{tab:ablation}(Row 2), when we run experiments without the context-query alignment component, the performance of our approach reduces to 67.4\%. This can be attributed to the fact that this component is responsible for the soft-alignment of the multimodal context and the query and without it the network struggles to locate the relevant answer of the question in the context.
\subsection{Emotion and Sentiment Understanding}
We observed that Social-IQ has many questions and answers that contain emotion or sentiment information. For instance, in Figure \ref{fig:qualitative}, the answer choices contain sentiment-oriented phrases like \textit{happy}, \textit{despise}, \textit{not happy}, \textit{not comfortable}. We performed an interesting study and divided the questions and answers in two disjoint sets. One set contained all the questions and answers which have emotion/sentiment-oriented words and the others which did not. We observe that both Tensor-MFN and the MCQA performed better on the set without emotion/sentiment-oriented words.

\section{Conclusion, Limitations and Future Work}
We present MCQA, a learning-based multimodal question answering task and evaluate our method on the Social-IQ, a benchmark dataset. We use co-attention for fusing and aligning the three modalities which serve as a context to the query. We  also use co-attention to also perform a context-query alignment to enable our network to focus on  the relevant parts of the video helpful for answering the query. We propose the use of a co-attention mechanism to handle combination of different modalities and question-answering alignment. It is a critical component of our model, as reflected by the drop in accuracy in the ablation experiments. While vanilla attention has been used for many NLP tasks, co-attention has not used for multimodal question answering; we clearly demonstrate its benefits. In practice, MCQA has certain limitations as it confuses and fails to predict the right answer multiple times. Social-IQ is a dataset containing videos of social situations and questions that require causal reasoning. We would like to explore better methods that can capture this reasoning. Our analysis in Section 4.5 suggests that the current models for question answering lack the understanding of emotions and sentiments and can be challenging problem in a question-answering setup. As part of future work, We would like to explicitly model this information using  multimodal emotion recognition techniques~\cite{mittal2019m3er,kim2018multimodal,majumder2018multimodal} to improve the performance of question answering systems.

\bibliography{acl2020}
\bibliographystyle{acl_natbib}
\end{document}